# A Semantic Communication System for Real-time 3D Reconstruction Tasks

1st Jiaxing Zhang
Department computer science and technology
Nanjing University of Aeronautics and Astronautics
Nanjing, China
jiaxinzhang@nuaa.edu.cn

2nd Luosong Guo
Department computer science and technology
Nanjing University of Aeronautics and Astronautics
Nanjing, China
luosongguo@nuaa.edu.cn

3rd Kun Zhu*
Department computer science and technology
Nanjing University of Aeronautics and Astronautics
Nanjing, China
zhukun@nuaa.edu.cn

4th Houming Qiu
Department computer science and technology
Nanjing University of Aeronautics and Astronautics
Nanjing, China
hmqiu56@nuaa.edu.cn

*Abstract*—Recently, 3D semantic maps have played an increasingly important role in high-precision robot localization and scene understanding. However, real-time construction of semantic maps requires mobile edge devices with extremely high computing power, which are expensive and limit the widespread application of semantic mapping. In order to address this limitation, inspired by cloud-edge collaborative computing and the high transmission efficiency of semantic communication, this paper proposes a method to achieve real-time semantic mapping tasks with limited-resource mobile devices. Specifically, we design an encoding-decoding semantic communication framework for real-time semantic mapping tasks under limited-resource situations. In addition, considering the impact of different channel conditions on communication, this paper designs a module based on the attention mechanism to achieve stable data transmission under various channel conditions. In terms of simulation experiments, based on the TUM dataset, it was verified that the system has an error of less than 0.1% compared to the ground-truth in mapping and localization accuracy and is superior to some novel semantic communication algorithms in real-time performance and channel adaptation. Besides, we implement a prototype system to verify the effectiveness of the proposed framework and designed module in real indoor scenarios. The results show that our system can complete real-time semantic mapping tasks for common indoor objects (chairs, computers, people, etc.) with a limited-resource device, and the mapping update time is less than 1 second.

## I. INTRODUCTION

Recently, the 3D semantic map has been widely applied in various robot applications. It can provide more precise localization accuracy and scene understanding, helping robots meet high-level task requirements[1]. Typically, the construction of a real-time semantic map requires a large number of resources and flexible mobility like high computational edge devices. However, these devices are expensive, restricting their widespread application[2]. Therefore, in order to reduce costs and enhance applicability, studying how to use cheap edge nodes with limited computational capabilities to achieve real-time semantic mapping holds significant research value.

To achieve good mobility and sufficient computing support, using cloud-edge collaborative computation, which transmits data from low computational edge nodes to high computational cloud servers for computing, is an effective solution[3]. In this setting, data should be transmitted as effectively as possible to achieve excellent real-time performance. In terms of effective data transmission, semantic communication is a new method that attracts widespread attention[4]. It extracts the semantic information from data and only transmits semantic features, which greatly reduces the amount of data transmission and achieves better real-time performance compared with traditional communication methods. Additionally, semantic communication can adapt to various channel conditions, achieving stable data transmission in all kinds of channel situations[5]. Currently, semantic communication is widely used in various types of data transmission and is combined with a range of tasks [6-7].

In terms of real-time semantic mapping tasks, many researchers have conducted research. Lai et al. propose a real-time semantic map construction system based on visual SLAM and CNN network[8]. Yang et al. propose a real-time semantic mapping method, achieving dense 3D semantic mapping with an RGB-D camera[9]. However, these studies only focus on improving the mapping accuracy and real-time performance with sufficient resources, instead of considering how to complete mapping tasks when the resources of edge devices are limited.

In order to broaden the application scopes, inspired by cloud-edge collaborative computation research, we introduce cloud assistance into resource-limited real-time semantic mapping tasks. Nevertheless, cloud-edge collaborative computation researches typically use traditional communication, resulting in long communication time for its large transmission volume and transmission errors in bad channel conditions[10].

These issues in traditional communication have been well considered in the research of semantic communication. In [11], it proposes an image semantic communication, which can compress the transmission data largely and achieve excellent real-time performance. In [12], it proposes a framework to transmit semantic segmentation map images over communication channels with different signal-to-noise ratios. Based on the aforementioned works, this paper proposes a semantic communication framework for resource-limited semantic mapping tasks. To the best of our knowledge, this is the first work proposed for the integration of cloud-edge collaborative computation into real-time semantic mapping tasks.

Specifically, the main contributions of this paper are as below:
- We propose an encoding and decoding semantic communication framework for semantic mapping tasks, achieving real-time completion of mapping tasks with limited-resource edge nodes.
- We design an attention module to adapt to different channel conditions, ensuring real-time and correct completion of semantic mapping tasks under various channel conditions.
- Considering the scarcity of prototype systems, we use a low computing edge node and a high computing cloud server to construct a cloud-edge collaborative prototype system for semantic mapping tasks, which effectively proves the validity of our proposed framework.

## II. SYSTEM MODEL

In our system, we consider two types of devices: the transmitter represented by edge devices, which have constrained computational resources but good mobility, and the receiver represented by cloud server, which has high computational capabilities and can generate semantic maps in real-time. As shown in Fig. 1, the communication process can be summarized into the following steps:

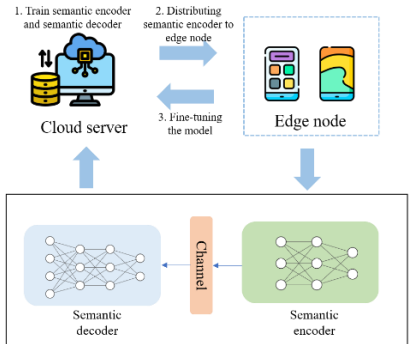

**Fig. 1** System scenario diagram.

1) The cloud server trains the encoder-decoder model for semantic communication and subsequently distributes the semantic encoder model to all edge devices through wireless transmission.

2) Edge devices conduct model fine-tuning based on real-world scenarios and extract semantic features, which will be transmitted to the cloud server for semantic recovery.

3) The cloud server performs data recovery based on the received semantic in-formation and completes semantic mapping tasks.

The edge devices are responsible for image acquisition and semantic feature extraction, while the cloud server is responsible for recovering semantic information and performing semantic mapping tasks based on the acquired semantic features. The simplified process of the system is depicted in Fig. 2.

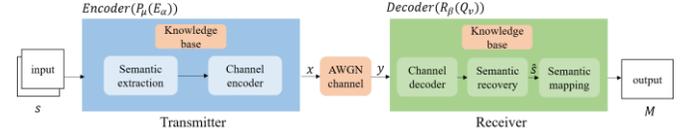

**Fig. 2** Simplified system framework diagram.

In this system, the transmitter is responsible for extracting semantic features and transmitting them through the channel. This process mainly consists of two parts: semantic extraction and channel encoder. Assuming the image source data is $s$, the overall transmission process can be represented as:
$$x = P_\mu(E_\alpha(s)), \quad (1)$$
where $E_\alpha()$ represents the semantic extraction network with training parameter $\alpha$, $P_\mu()$ refers to the channel encoder network with training parameter $\mu$. The final output $x$ is subsequently sent to the channel for transmission.

In order to account for channel fluctuations and achieve channel variability, we select the Adaptive White Noise Channel (AWGN) as the system channel, denoted by Eq. (2):
$$y = x + n, \quad (2)$$
where $n$ denotes the channel noise, which follows a normal distribution $N(0, \sigma^2)$ and $y$ represents the channel output, which will be transmitted to the receiver for subsequent steps. The receiver is responsible for recovering semantic information and constructing semantic maps. This process mainly includes three parts: channel decoder, semantic recovery, and semantic mapping. As shown in Eq. (3), the semantic recovery process is the inverse procession of the semantic feature extraction.
$$\hat{s} = R_\beta(Q_\nu(y)), \quad (3)$$
where $Q_\nu()$ denotes the channel decoder network with training parameter $\nu$, $R_\beta()$ denotes the semantic decoder network with training parameter $\beta$. The output of semantic recovery $\hat{s}$ will serve as the input for subsequent semantic mapping. For semantic mapping, Semantic Simultaneous Localization and Mapping (Semantic SLAM) is used to generate the semantic map $M$ in real time.

The knowledge base serves as the fundamental component of the semantic communication mapping system. Shared background knowledge between the transmitter and receiver helps to comprehend and interpret the exchanged information. In this system, the knowledge base is represented by the structure and parameters of neural networks.

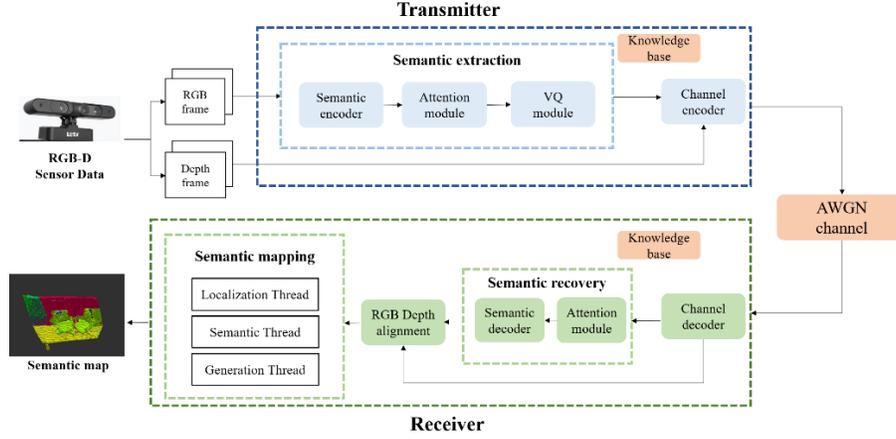

**Fig.3** The encoding-decoding framework of semantic communication mapping system.

### III. PROPOSED FRAMEWORK

In this section, we will mainly introduce the architecture of our proposed semantic communication mapping system, as shown in Fig. 3. Firstly, a detailed overview is provided to introduce the various components within the proposed framework. Subsequently, some design details of the attention module and the network structures of the transmitter and receiver are discussed.

*A. The Architecture of System Framework*

The overall system consists of an encoding-decoding structure. At the encoding end, the RGB-D sensor is chosen as the data acquisition sensor, which captures both RGB frames and depth frames simultaneously. The RGB frames will be fed into the semantic extraction module, while depth frames will be directly fed into the channel encoder module due to its lack of semantic information. The semantic extraction module consists of three components: semantic encoder, attention module, and vector quantization module. In the semantic encoder, a Convolutional Neural Network (CNN) based method is adopted to extract semantic features. After obtaining semantic features, an attention module is added to adapt to various channel conditions. This attention module is designed based on the perceived signal-to-noise ratio. More details about the attention module will be introduced in the following section.

In order to further compress semantic information, we specially design a vector quantization module. The vector quantization module maps continuous information to finite-dimensional discrete vectors, which can achieve maximum compression of semantic data without corrupting them. The quantized compressed data will be concatenated with the depth frames and then sent to the channel by the channel encoder.

At the receiver end, the channel decoder is first performed after receiving the signal. It helps restore the signal to its previous semantic information. The recovery process of semantic information is similar to the reverse process of the previous transmission process. Specifically, it includes two parts: the attention module and the semantic decoder based on CNN. The ultimate goal is to restore semantic feature information to the original RGB frames. Compared to the complex process of RGB frames, depth frames are directly transmitted and are faster than RGB frames. Therefore, when RGB frames are restored, there is misalignment caused by the time difference between them. In order to address the misalignment problem, a frame alignment module is specially designed to match RGB frames and depth frames based on preset timestamps. Only successfully aligned image frame pairs will be sent to the subsequent semantic mapping.

For semantic mapping, high mapping accuracy and good real-time performance are our ultimate goals. There are three parts to this module: localization, semantic extraction, and semantic map generation. To achieve good real-time performance, these three modules are deployed in three threads for parallel execution.

In this encoding and decoding architecture, the encoding end only needs to extract semantic information, requiring minimal computational capacities. Meanwhile, the amount of quantified semantic information is very small, ensuring good real-time performance and real-time semantic mapping tasks will be completed at the decoding end.

*B. Attention Module Design*

In neural networks, attention is a mechanism that reallocates resources that are originally evenly distributed according to the importance of attention objects. Conventional attention mechanism consists of channel attention and spatial attention, which focus on the input of different channel information and the input of different location information, respectively. However, both of them do not focus on resource allocation under different channel conditions. Therefore, we design the attention module that is based on signal-to-noise ratio perception and automatically learns the corresponding attention weights for channels with different signal-to-noise ratios through training models.

As shown in Fig. 4, a fully connected layer is utilized with the signal-to-noise ratio (SNR) as input, and a sigmoid function is used to obtain attention weights in the attention module. Then the attention weights are expanded to match the shape of input features, which will then be multiplied with the input features.

The attention layer can be represented as
$$z_{f+1} = f_{FCA}(z_f, snr) * z_f, \quad (4)$$
where $z_f$ and $z_{f+1}$ denote the input and output features respectively. $f_{FCA}()$ is a fully connected layer followed by a sigmoid function.

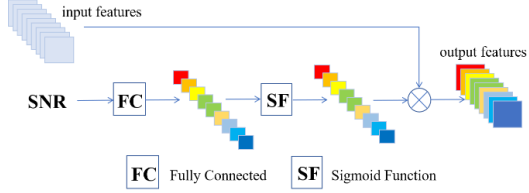

**Fig. 4** Attention module structure design.

*C. Transmitter*

The role of the transmitter is to extract semantic information from the source data, mainly including semantic encoder, attention module, and vector quantization. As shown in Fig. 5, almost all of them are implemented by neural networks. The semantic encoder extracts the semantic features from the RGB frames $S \in R^{H \times W \times 3}$ through two kinds of down-sampling modules, where $H$, $W$, and 3 are image width, image height, and the number of channels. Specifically, the semantic encoder module comprises two convolutional down-sampling modules and two residual down-sampling modules. Each convolutional down-sampling module includes a convolutional layer and a BatchNormalization layer. Between two modules, an activation function ReLU is added to increase the nonlinear capability. After the convolutional down-sampling operations, two residual down-sampling modules, which include a residual layer based on ResNet and BatchNormalization layer. Thus, the semantic features are extracted.

After semantic encoding, the semantic features are sent into the attention module to add attention weights. The output feature tensor of the attention module enters the vector quantization part to compress the data volume. This idea of vector quantization is that the continuous input data will find the closest discrete semantic embedding vector to ensure the transmitter only send these semantic vectors instead of the whole semantic features, greatly reducing the amount of data transmission[13]. Specifically, the module will search for the nearest vector in the semantic embedding space, as shown in Eq. (5):
$$n = arg_n \, min\|f_m - e_n\|_2, \quad (5)$$
where $f_m$ is the semantic encoding result, $e_n$ is the semantic embedding space shared by the transmitter and receiver. After vector quantization, the discrete semantic features are inputted to the channel encoder together with depth frames and then sent to the channel for transmission.

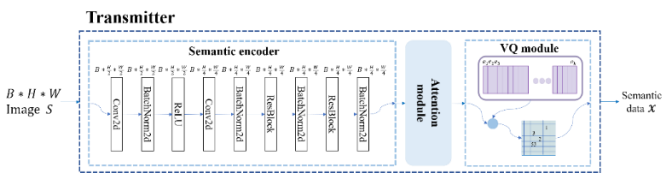

**Fig. 5** Transmitter network structure diagram.

*D. Receiver*

The role of the receiver is to recover the extracted semantic information and use the recovery results for semantic mapping. Its network is shown in Fig. 6, which can be divided into three components: attention module, semantic decoder, and semantic mapping. After passing through the attention module, semantic features are restored to the original image by the semantic decoder. In network design, the components of decoder networks are similar to encoder networks while having inverse functions. The up-sampling part shown in Fig. 6 is successively connected by the deconvolution layer, residual block, and BatchNormlization layer. Like semantic encoder, the role of ReLU is to increase nonlinearity. After deconvolution and upsampling, the image will be restored by changing the result to [0,1] through tanh.

In addition to the semantic recovery part, the receiver also includes the semantic mapping part. To ensure the real-time operation of the system, three threads are used in parallel. For the localization thread, we designed based on ORB-SLAM3 [14] while removing the mapping part and retaining only self-localization and loop closure detection. For the semantic extraction module, semantic segmentation is chosen for obtaining high-precision semantic information. Besides, considering the real-time requirements of the system, it is necessary to complete the semantic segmentation task as quickly as possible. Therefore, we choose SCTNet[15], whose structure separates training and inference effectively to ensure real-time performance, as the semantic segmentation network. After obtaining semantic information, this semantic mapping module will construct the semantic point cloud based on the depth information and the semantic information. In order to better integrate semantic information and point clouds, we choose Bayesian fusion to achieve high-precision semantic point cloud fusion. Finally, the semantic map, whose form is octree map for saving storage space and improving mapping efficiency, is generated.

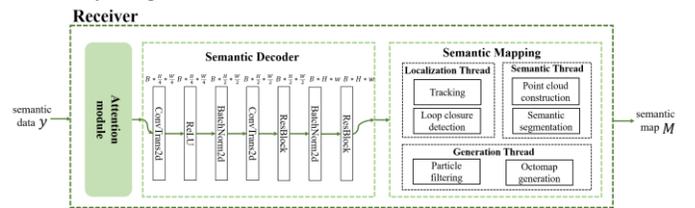

**Fig. 6** Receiver network structure diagram.

IV. EVALUATION AND RESULT

To verify the effectiveness of the proposed framework and designed modules for the limited-resource semantic mapping tasks, several experiments are designed in this section. Firstly, several simulation experiments are conducted to demonstrate its superiority on semantic mapping tasks compared with other communication algorithm. Secondly, an ablation study is proposed to validate the effectiveness of the attention module we propose. Finally, a prototype system is implemented in real scenarios to visually demonstrate the performance of the system.

## A. Simulation Validation on System Performance

*a) Real-time Performance Validation:* For semantic mapping tasks, real-time performance is a key metric. In our system, the semantic features are extracted by the transmitter from the input image, and then restored by the receiver for semantic mapping tasks. Here, we utilize the processing time of images with different commonly used resolutions to evaluate the real-time performance. Specifically, we use a stable AWGN channel as the test channel and use an RGB-D sensor to pre-capture input data sources with pre-designed size. In order to demonstrate the real-time advantages of our system, we choose some baseline algorithms (JPEG, VQ-VAE[13], Deep-JSCC[16]) for comparison. As depicted in Fig. 7, it is evident that our real-time performance surpasses JPEG and Deep-JSCC under several specific image sizes, particularly in transmitting large volumes of data. Regarding VQ-VAE, which similarly utilizes vector quantization, we achieve comparable real-time performance while significantly enhancing transmission quality, as introduced in the ablation study below.

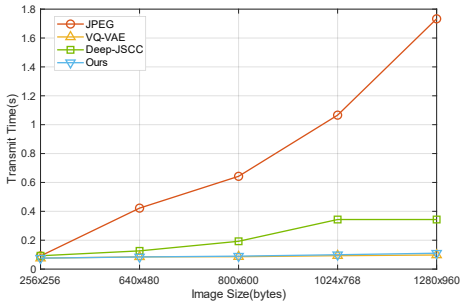

**Fig. 7** Real time performance comparison of with different data sizes.

*b) Semantic Mapping Accuracy Validation:* Typically, the established semantic map effect and the localization accuracy during the mapping process are widely used in semantic mapping research to measure the mapping accuracy [8]. To verify the mapping accuracy of our system, we choose the Tum dataset freiburg2_pioneer_slam2 as the source data. Freiburg2_pioneer_slam2 is an indoor dataset with RGB-D sensor data, and provides ground-truth to help correct comparison. The result of semantic map construction is shown in Fig. 8. It can be observed that, in comparison to other baseline algorithms, our method offers a more comprehensive map construction, particularly in terms of capturing the entire outline of the house and generates more map content at the same time. Notably, there are some purple spots on the yellow ground. This is because there are many messy wires on the ground, and the semantic segmentation algorithm distinguishes it from the empty ground.

*c) Localization Accuracy Validation:* For localization accuracy, as mentioned above, the localization part in the receiver is based on ORB_SLAM3 which has a great localization effect. Therefore, we choose ORB_SLAM3 directly as the ground-truth to measure the localization accuracy. As for the environment, freiburg2_pioneer_slam2 is used again. To achieve more quantitative comparison, some quantitative indexes are selected as localization error indicators: Absolute Trajectory Err or (ATE) and Relative Pose Error (RPE)

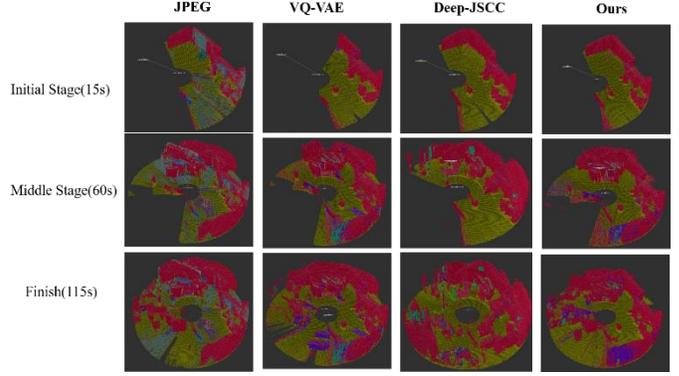

**Fig. 8** Semantic mapping effect using Tum dataset.
(Red: wall, Yellow: empty ground, Purple: ground with debris)

As shown in Table 1, our method achieves the minimum error on both indicators, with the differences between them being less than 0.1%, proving our system has great localization effect.

TABLE I. THE LOCALIZATION ERROR INDICATORS COMPARED WITH GROUND-TRUTH

| Algorithm | Localization error indicators | |
|---|---|---|
| | ATE↓ | RPE↓ |
| Deep-JSCC | 0.000115 | 0.000821 |
| VQ-VAE | 0.000235 | 0.000295 |
| JPEG | 0.000178 | 0.000492 |
| Ours | **0.000068** | **0.000097** |

## B. Ablation Study

Unstable channels are often an important factor affecting communication efficiency. To verify the effectiveness of the designed attention module under different channel conditions, we conduct an ablation study in the simulation environment. We adopt the AWGN channel, which controls the noise intensity by adjusting the signal-to-noise ratio（SNR）. The higher SNR, the smaller the noise. In terms of the metrics, we use peak signal-to-noise ratio (PSNR), where higher values indicate better transmission performance. The results are shown in Fig. 9, our method with the attention module performs best performance under various SNR conditions. In addition, compared to other algorithms, the method with attention module has smaller fluctuations under all channel conditions, which proves better adaptability to various channel conditions.

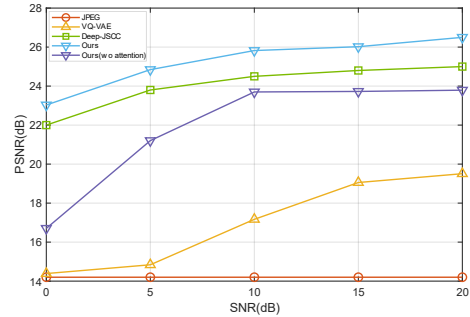

**Fig. 9** Comparison of transmission quality under different SNR.

## C. Real Scene Prototype System Implementation

In the implementation of the prototype system, according to the encoding-decoding framework proposed in this paper, the encoding end is a resource-limited edge device, and the decoding end is a high computing cloud server. Specifically, as Fig. 10 shows, a robot with Nvidia Jetson TX2 is used as the transmitter for encoding. The resources of the edge device are not sufficient to complete real-time semantic 3D mapping tasks, but it is sufficient to extract semantic information. The receiver is the host server for decoding, with an AMD EPYC 7B12 64-Core CPU and Nvidia GFX 3090 GPU. The high computational capacities receiver can complete the semantic recovery and semantic mapping tasks in real time.

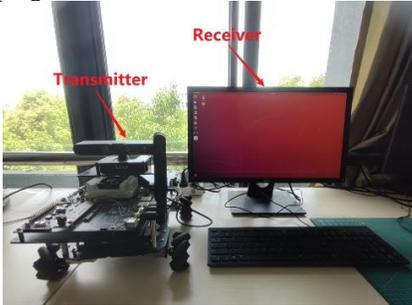

**Fig, 10** Realistic semantic communication mapping scenarios.

For the RGB-D sensor, we chose the Xtion, which can provide the RGB frames and depth frames simultaneously. In this paper, the indoor environment is the focus of our attention. Therefore, we conduct some common used objects (chairs, computers, people, etc.) in real scenarios. As shown in Fig.11, the semantic communication mapping system can finish the mapping task in real-time while the traditional way fails to construct in the limited resources.

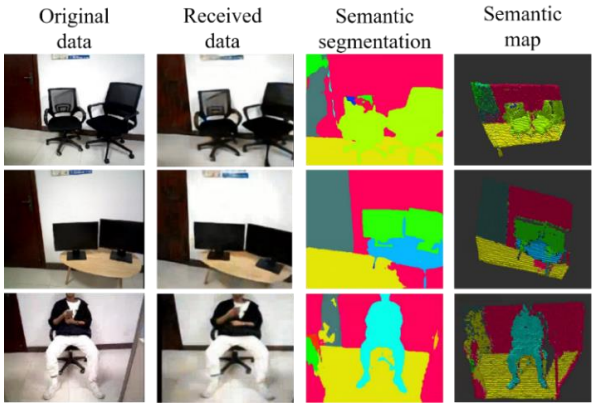

**Fig. 11** The semantic mapping of indoor objects
(Pale yellow: chair, Green: computer, Light blue: person).

## V. CONCLUSION

This paper focuses on real-time semantic mapping tasks with low computing edge devices. By integrating the cloud-edge collaborative computation and semantic communication, an encoding-decoding framework is proposed to achieve real-time semantic mapping tasks under limited resources situations. Considering the possibility of unstable communication channels, an attention module is specifically designed to ensure effective data transmission under various channel conditions. Numerous simulation experiments and prototype system in real scenarios have verified the effectiveness of the proposed framework in semantic mapping accuracy and real-time performance.

# Author's Background Information

**Note:** *Please fill in every author's background information. It will help us understand the paper better and **will not be published.**

*Title could be Prof., Assoc. Prof., Assist. Prof., Dr., Senior Lecturer, etc.

| Author's name: | Position: | Research areas: | Personal Website: |
|---|---|---|---|
| Jiaxin Zhang | Dr. | Semantic communication | |
| Luosong Guo | Dr. | Semantic communication, 3D reconstruction | https://scholar.google.com/citations?hl=en&user=PMP123IAAAAJ |
| Kun Zhu | Prof. | Resource management, wireless network | https://scholar.google.com/citations?user=XRLvEPYAAAAJ&hl=en&oi=ao |
| Houming Qiu | Dr. | Edge computing, coded distributed computing | https://scholar.google.com/citations?hl=en&user=4UXLAawAAAAJ |